\documentclass[sigconf]{acmart}

\AtBeginDocument{%
  }

\setcopyright{cc}
\setcctype{by}
\copyrightyear{2026}
\acmYear{2026}
\acmDOI{XXXXXXX.XXXXXXX}
\acmConference[KDD '27]{Proceedings of the 33rd ACM SIGKDD Conference on Knowledge Discovery and Data Mining}{August, 2027}{San Jose, CA, USA}
\acmISBN{978-1-4503-XXXX-X/2027/08}

\usepackage{stfloats}
\usepackage{colortbl}
\usepackage{siunitx}
\usepackage{multirow}

\AtBeginDocument{\citestyle{acmnumeric}\setcitestyle{numbers,sort&compress}}

\usepackage{xcolor}
\definecolor{LinkGray}{RGB}{85,85,85}
\AtEndPreamble{\hypersetup{
    colorlinks=true,
    linkcolor=LinkGray,
    citecolor=LinkGray,
    urlcolor=LinkGray,
    filecolor=LinkGray,
    menucolor=LinkGray,
    anchorcolor=LinkGray}}
\makeatletter
\AtEndPreamble{
    \patchcmd{\ACM@mk@linecount}{\color{red}}{\color{LinkGray}}{}{}
    \patchcmd{\ACM@mk@linecount}{\color{red}}{\color{LinkGray}}{}{}
}
\makeatother

\newcommand{\dsym}{\ensuremath{\Delta}}
\newcommand{\EddyFlow}{\textsf{EddyFlow}}

\begin{document}

\title[Mesoscale-Preserving SST Downscaling Across Ocean Basins]{Transferable Dual-Stream Representations for Mesoscale-\\Preserving Sea Surface Temperature Downscaling}

\author{Parth Doshi}
\orcid{0009-0009-6359-8330}
\affiliation{%
    \institution{Dalhousie University}
    \city{Halifax}
    \state{NS}
    \country{Canada}
}
\email{parth.doshi@dal.ca}

\author{Priyanka Aravindan}
\orcid{XXXX-XXXX-XXXX-XXXX}
\affiliation{%
    \institution{Dalhousie University}
    \city{Halifax}
    \state{NS}
    \country{Canada}
}
\email{priyanka.aravindan@dal.ca}

\author{Vaishnav Vaidheeswaran}
\orcid{XXXX-XXXX-XXXX-XXXX}
\affiliation{%
    \institution{Dalhousie University}
    \city{Halifax}
    \state{NS}
    \country{Canada}
}
\email{vaishnav@dal.ca}

\author{Md Mahbub Alam}
\orcid{0000-0003-0756-264X}
\affiliation{%
    \institution{Dalhousie University}
    \city{Halifax}
    \state{NS}
    \country{Canada}
}
\email{mahbub.alam@dal.ca}

\author{Gabriel Spadon}
\authornote{Corresponding Author}
\orcid{0000-0001-8437-4349}
\affiliation{%
    \institution{Dalhousie University}
    \city{Halifax}
    \state{NS}
    \country{Canada}
}
\email{spadon@dal.ca}

\renewcommand{\shortauthors}{Doshi et al.}

\begin{abstract}
    Deep learning models for scientific spatio-temporal downscaling often minimize reconstruction error while failing to preserve physically meaningful multi-scale structure. For sea surface temperature prediction, this can yield outputs that are numerically plausible yet overly smooth, missing mesoscale variability critical to regional ocean dynamics. Existing methods often focus on pixel-wise objectives or single-context conditioning, which limits their ability to preserve spectral fidelity and generalize across regions. To address this, we propose \EddyFlow{}, a representation learning framework for kilometer-scale sea surface temperature downscaling that balances predictive accuracy, scale-dependent structure, and regional generalization. \EddyFlow{} is trained on the Gulf of St.~Lawrence and evaluated in zero-shot and few-shot settings on the Bay of Fundy and the Gulf of Mexico. \EddyFlow{} demonstrates that physics-informed representation learning reduces zero-shot RMSE by 21\%, achieves up to 85.6\% skill relative to persistence on unseen domains, and maintains near-ideal spectral fidelity with a PSD ratio of $\approx 1.00$.
\end{abstract}



\received{26 July 2026}

\maketitle

\section{Introduction}
\label{sec:intro}

Sea Surface Temperature (SST) is a primary driver of coastal ecosystem health, fisheries productivity, and marine weather hazards~\cite{STOCK2015219}. However, satellite and reanalysis datasets that provide global SST estimates are far too coarse to capture the frontal structures, eddies, and thermal gradients that govern regional ocean dynamics~\cite{sst_regime_shifts_2026,saxena2021efficient,thiria2023downscaling,multisource_sst_gan}. In the Gulf of St.~Lawrence (GSL), sharp SST fronts associated with the Laurentian Channel and shelf-break bathymetry shape fish habitats and marine management decisions, yet remain unresolved in standard reanalyses such as ERA5\footnote{\url{https://cds.climate.copernicus.eu/datasets/reanalysis-era5-single-levels}}~\cite{gsl_biogeochem_2021,era5}. Recovering kilometer-scale SST from coarse inputs is a prerequisite for fisheries management and coastal hazard forecasting, which depend on precise thermal gradients, not approximate magnitudes~\cite{gsl_biogeochem_2021,sst_regime_shifts_2026}.

Deep learning models for structured prediction increasingly infer high-resolution scientific fields from sparse or coarse observations~\cite{saxena2021efficient, thiria2023downscaling, multisource_sst_gan, lupin2025regional_emulator}. Even at strong predictive accuracy, pixel-wise losses alone do not guarantee that learned representations capture target-domain structure. Representations here mean the latent encoding governing how spatial organization, temporal evolution, and cross-scale dependencies are preserved. When these encodings miss physical structures, models can achieve low reconstruction error while producing outputs that are spectrally smooth, temporally uninformative, or unable to generalize across physical regimes, a failure often visible only beyond the training distribution~\cite{thiria2023downscaling,lupin2025regional_emulator}.

SST downscaling exemplifies this limitation as the target application reflects coupled physical processes, so faithful reconstruction requires more than interpolating the coarse input~\cite{lupin2025regional_emulator}. Models trained to minimize pixel-wise reconstruction error can match target values while collapsing fine-scale variance~\cite{multisource_sst_gan,thiria2023downscaling}. This collapse goes unpenalized by pixel-wise losses, yielding compressed predictions structurally inconsistent with the dynamics and fragile under geographic shift~\cite{lupin2025regional_emulator}.

The difficulty is sharpest where spatial statistics are not stationary and temporal persistence is strong. The GSL exemplifies both. Its semi-enclosed geometry, freshwater input from major rivers, and strong coupling to shelf circulation produce an SST field whose spatial covariance structure varies across the domain. Consequently, the mapping from coarse to fine resolution is non-stationary, forcing region-dependent relationships rather than a global interpolation rule. Convolutional or attention-based architectures~\cite{mardani2024corrdiff, diffds2024} often assume spatial homogeneity and struggle to generalize. Temporal autocorrelation makes persistence a competitive baseline.

These observations suggest viewing SST downscaling not merely as image super-resolution, but as representation learning under domain shift. The learned encoding must remain accurate, temporally informative, and spectrally faithful when the physical regime changes across geographies. We assess these properties with three metrics. Root-mean-square Error (RMSE) measures pixel-wise reconstruction error, the Skill Score (SS) measures improvement over the ERA5 persistence baseline, and the Power Spectral Density (PSD) ratio measures preservation of fine-scale variability. No single metric captures all three; their joint behavior determines whether learned representations are accurate, informative, and physically consistent.

We present \EddyFlow{}, a physics-informed representation-learning framework for kilometer-scale SST downscaling, evaluated under geographic domain shift. We adopt the GSL as the training domain because its physical complexity, non-stationary spatial statistics, and strong persistence test whether a model learns generalizable structure rather than dataset-specific patterns (Section~\ref{sec:problem}). \EddyFlow{} separates atmospheric forcing, ocean memory, and diffusion-based refinement to recover the mesoscale variance that pixel-wise objectives smooth away (Section~\ref{sec:method}), mesoscale means intermediate size or scale, typically ranging from roughly 2 to 1,000 kilometers. \EddyFlow{} is evaluated in zero-shot and few-shot transfer on the Bay of Fundy (BOF) and the Gulf of Mexico (GOM) in Section~\ref{sec:experiments}. Physics-informed representation learning drives generalization across ocean regimes; our contributions:
\begin{itemize}
    \item a \textbf{21\% reduction in zero-shot RMSE} on unseen basins relative to a strong downscaling baseline;
    \item up to \textbf{85.6\% skill relative to the ERA5 persistence baseline} on domains never observed during training; and
    \item \textbf{near-ideal spectral fidelity}, with a PSD ratio of $\approx 1.00$ that preserves fine-scale variability rather than blurring it.
\end{itemize}
The remainder of the paper is organized as follows. Section~\ref{sec:related} reviews related work; Section~\ref{sec:problem} formalizes the downscaling task and evaluation criteria; Section~\ref{sec:method} presents \EddyFlow{}; Section~\ref{sec:experiments} reports in-domain and transfer results; and Section~\ref{sec:conclusion} concludes.

\section{Related Work}
\label{sec:related}

\textit{\textbf{Learning high-resolution spatial structure.}}
Much recent work treats geophysical downscaling as a spatial reconstruction problem, in which diffusion and generative models recover the sharp gradients and high-frequency variability that pixel-wise regression smooths away~\cite{wassdiff2024,consistencydownscale2025}. CorrDiff~\cite{mardani2024corrdiff} pairs a deterministic backbone with a residual diffusion corrector, restricting stochastic generation to unresolved fine scales, improving realism while preserving spectral behavior. DIFFDS~\cite{diffds2024} adapts diffusion-based image restoration to SST, learning a compact high-resolution prior that guides reconstruction from coarse inputs.
Earlier work established that deep networks can recover ocean-front structure from satellite imagery. Lloyd et al.~\cite{lloyd2021optical} fuse optical and thermal features for fivefold SST super-resolution, and Ducournau and Fablet~\cite{ducournau2016superresolution} showed that CNN-based super-resolution substantially outperforms classical downscaling methods. Recent Mediterranean studies with deterministic and generative formulations~\cite{mediterranean_sst_sr_2024, sst_gan_2025} reinforce that learning a conditional high-resolution mapping from low-resolution SST alone, outperforms interpolation, particularly in recovering small-scale gradients and spectral properties. These methods excel at spatial fidelity and high-wavenumber variance recovery, but most refine one or a few concurrent observations with limited temporal conditioning, so improved visual and spectral quality does not guarantee evolution consistent with recent atmospheric forcing.As they are largely translation-equivariant architectures, they struggle to capture the fixed, spatially non-stationary influence of coastlines, bathymetry, and continental shelves.

\noindent\textit{\textbf{Learning temporal ocean evolution.}}
A second line of work treats ocean prediction as sequence modeling, learning state evolution through autoregressive or multi-step temporal models that better represent persistence, advection, and slowly varying circulation structure~\cite{krestenitis2023sst,stacked2024sst}. WenHai~\cite{WenHai} shows that explicit temporal modeling is essential for maintaining coherent evolution in autoregressive settings, where small stepwise errors otherwise accumulate. GraphCast~\cite{lam2023graphcast} and GenCast~\cite{price2024gencast} demonstrate that graph- and diffusion-based temporal emulators achieve strong forecast skill while preserving dynamical coherence across lead times, and FuXi-Ocean~\cite{fuxiocean2026} extends this to sub-daily, eddy-resolving global forecasting with improved skill over WenHai. Forecasting, however, solves a different problem from SST downscaling. It predicts future ocean states rather than recovering fine-scale structure missing from coarse observations, capturing large-scale evolution while missing sharp fronts and kilometer-scale variability. Many forecasting models also encode atmospheric forcing and ocean memory in a single latent representation, limiting separate modeling of short-term dynamics from persistent ocean structure observations.

\noindent\textit{\textbf{Learning regional generalization.}}
A third line of work emphasizes that ocean prediction models must remain reliable when transferred across water bodies with distinct coastlines, boundary conditions, and circulation regimes, since one coarse input can map to different fine-scale responses depending on local geometry and context. Recent regional emulators combine low-resolution autoregressive forecasting with learned high-resolution refinement and online bias correction, remaining stable over long horizons while improving spectral behavior and fine-scale realism~\cite{lupin2025regional_emulator}, showing that physical consistency, roll-out stability, and multiscale correction can coexist in one framework. Such models, however, are typically trained on a single basin, and their performance degrades substantially when out of the training distribution, making them effective basin-specific models but unreliable on unseen regions.

\section{Problem Formulation}
\label{sec:problem}

Global SST products such as ERA5 capture the large-scale thermal structure of the ocean but are too coarse to resolve fine-scale features such as fronts, eddies, and coastal gradients. Statistical downscaling therefore aims to recover a high-resolution SST field consistent with the observed coarse state~\cite{downscaling_review_2022}. This is an ill-posed problem because multiple plausible high-resolution fields can correspond to the same coarse input. As a result, models trained only with pixel-wise regression losses often predict overly smooth fields that minimize average error but fail to recover realistic fine-scale variability. Diffusion models address this by refining an initial estimate towards consistent fine-scale structure~\cite{ho2020ddpm,karras2022}.

Let $x_t$ denote the coarse-resolution input at time $t$, $o_t$ denote the optional ocean-state history, and $y_t$ denote the target high-resolution SST field. The objective is a mapping $f_{\theta}$ from coarse input and history to the high-resolution field $\hat{y}_t = f_{\theta}(x_t, o_t)$. Since SST evolves gradually, we estimate the temporal residual $r_t = y_t - y_{t-1}$ rather than the full field, and recover the target by adding it back to the previous day through $\hat{y}_t = y_{t-1} + \hat{r}_t$. Predicting the residual focuses the model on day-to-day spatial change rather than repeatedly reconstructing the slowly varying background at every timestep. It also yields a natural persistence reference that carries the previous day's state forward and thus assumes the SST remains unchanged absent new forcing.

These observations motivate the three complementary evaluation criteria used throughout this study: RMSE for pixel-wise accuracy, persistence-relative skill score for temporal informativeness, and PSD ratio for spectral fidelity. Beyond metrics, we evaluate in zero-shot and few-shot transfer settings. A model that performs well only within its training basin may simply have learned basin-specific coastline or circulation patterns rather than the coarse-to-fine relationship. Testing on unseen basins therefore better measures whether the representation generalizes across geographies.

\section{Methodology}
\label{sec:method}

\subsection{RMSE and the Skill Score}
\label{sec:ss}

SST demonstrates strong day-to-day persistence~\cite{bulgin2020}. The simplest predictor, $\hat{y}^{\mathrm{pers}}_t = y_{t-1}$, achieves low RMSE in most coastal and shelf regions because SST rarely changes significantly within a 24-hour period. This situation presents a methodological challenge where a model may report an RMSE that appears excellent in absolute terms while providing no additional predictive information, simply by replicating persistence. Consequently, RMSE is necessary but insufficient, as it cannot distinguish between a model that learned ocean dynamics and one that reproduces the previous day's field.

We therefore define a skill score relative to a persistence reference, following the classical verification framework in which a forecast is judged not against zero, but against a reference forecast requiring no learning component~\cite{cliper_persistence_1997}. Let $\mathrm{RMSE}(\hat{y}_t, y_t)$ denote the root-mean-square error of the model prediction and $\mathrm{RMSE}(\hat{y}^{\mathrm{pers}}_t, y_t)$ that of the persistence reference. We define the skill score as:
\begin{equation}
    \mathrm{SS} = 1 - \frac{\mathrm{RMSE}(\hat{y}_t, y_t)}{\mathrm{RMSE}(\hat{y}^{\mathrm{pers}}_t, y_t)}.
    \label{eq:skill}
\end{equation}

The ratio is directly interpretable. When $\mathrm{SS} = 0$, the model is exactly as informative as carrying the reference field forward. When $\mathrm{SS} > 0$, the model has extracted genuine predictive signal beyond persistence, and when $\mathrm{SS} < 0$, the model is \emph{worse} than the naive baseline despite an RMSE that looks reasonable in isolation. This last case is the one that a pixel-wise loss alone cannot detect, which is why persistence-relative skill, not RMSE alone, is the correct lens for short-horizon SST prediction. In our zero-shot BOF and GOM setting, a positive skill score is the minimum bar for indicating model usefulness. The skill score, however, is computed on raw pixel values and is blind to \emph{where} error lives across spatial scale. A model can satisfy $\mathrm{SS} > 0$ while discarding the fine-scale structure that makes it useful, a failure visible only in the frequency domain.

\subsection{Pixel-Wise Loss and the PSD Ratio}
\label{sec:psdr}

Minimizing squared error drives a network toward the conditional mean of the target at each pixel; when fine-scale structure is only partially determined by the coarse input, that mean is smoother than any single realization, so a pixel-wise reconstruction loss (the loss term $\mathcal{L}_{\mathrm{rec}}$ in Section~\ref{sec:loss}) alone yields pixel-accurate yet systematically under-variant predictions at high wavenumbers, a deviation we quantify in the frequency domain, scale by scale.

Let $y(\mathbf{u})$ be the target field and $\hat{y}(\mathbf{u})$ the prediction over spatial position $\mathbf{u}=(u_1,u_2)$, each with its spatial mean removed,
\begin{equation}
    y'(\mathbf{u}) = y(\mathbf{u}) - \mu_y.
\end{equation}
The 2D Fourier transform of the centered field and its inverse are
\begin{equation}
    Y(\mathbf{k}) = \iint y'(\mathbf{u})\,e^{-i\mathbf{k}\cdot\mathbf{u}}\,d\mathbf{u}, \quad y'(\mathbf{u}) = \tfrac{1}{(2\pi)^2}\!\iint Y(\mathbf{k})\,e^{i\mathbf{k}\cdot\mathbf{u}}\,d\mathbf{k},
\end{equation}
with wavenumber $\mathbf{k}=(k_x,k_y)$, and Parseval's theorem guarantees
\begin{equation}
    \iint |y'(\mathbf{u})|^2\,d\mathbf{u} = \frac{1}{(2\pi)^2} \iint |Y(\mathbf{k})|^2\,d\mathbf{k},
\end{equation}
so spatial variance is exactly preserved and attributable spectrally.

\noindent
The power spectra of the target field and the model prediction are
\begin{equation}
    P_y(\mathbf{k}) = |Y(\mathbf{k})|^2, \qquad P_{\hat{y}}(\mathbf{k}) = |\hat{Y}(\mathbf{k})|^2,
\end{equation}
where $\hat{Y}(\mathbf{k})$ denotes the transform of the centered prediction. Ocean fields are scale-organized rather than direction-organized; consequently we radially average over the annulus $\Omega_k$ at $k = \sqrt{k_x^2+k_y^2}$,%
\begin{equation}
    \bar{P}_y(k) = \frac{1}{|\Omega_k|} \sum_{\mathbf{k}\in\Omega_k} P_y(\mathbf{k}),
\end{equation}
and identically for $\bar{P}_{\hat{y}}(k)$, giving the isotropic 1D PSD used in downscaling and turbulence diagnostics~\cite{mardani2024corrdiff}. We define the PSD ratio as:
\begin{equation}
    \mathrm{PSDR}(k) = \frac{\bar{P}_{\hat{y}}(k)}{\bar{P}_y(k) + \varepsilon},
\end{equation}
with $\varepsilon$ a small constant for numerical stability, evaluated between the coarse-input scale and the kilometer-scale Nyquist limit, reported over 5--50~km (Section~\ref{sec:experiments}). A value of $\mathrm{PSDR}(k)\approx 1$ indicates that variance is preserved at scale $k$, while $\mathrm{PSDR}(k)<1$ indicates over-smoothing, the spectral signature of mean-seeking collapse, and $\mathrm{PSDR}(k)>1$ indicates over-amplified noise. A model can post low RMSE and $\mathrm{SS}>0$ while $\mathrm{PSDR}(k)\to 0$ at high $k$, exactly the failure mode invisible to pixel-wise metrics, since eddies and fronts occupy bands of $k$; a faithful model must not collapse.

\subsection{Two Streams for Two Timescales}
\label{sec:streams}

Atmospheric forcing and oceanic memory operate on separable timescales. Synoptic weather systems driving short-term SST change have lifetimes of two to seven days, while SST anomalies set by that forcing persist for weeks to months as the mixed layer relaxes toward equilibrium~\cite{bulgin2020}. One encoder at one window and sampling rate cannot resolve both, so \EddyFlow{} has two input tensors,
\begin{equation}
    X_a \in \mathbb{R}^{28 \times 23 \times 20 \times 48}, \qquad X_o \in \mathbb{R}^{60 \times 1 \times 501 \times 1201},
\end{equation}
with dimensions ordered as frames, channels, height, and width. $X_a$ is a 7-day, 6-hourly window of 23 coarse atmospheric reanalysis forcing fields, spanning roughly two synoptic cycles, beyond which autocorrelation decays too quickly to add information. $X_o$ is a 60-day, daily window of fine-resolution ocean surface temperature observations, chosen so that mesoscale eddies translating at 5--20 km/day~\cite{chelton2011eddies} remain inside the receptive field even after several hundred kilometers, and long enough to span seasonal transitions such as summer heating giving way to autumn cooling.

Because the two streams differ in resolution by roughly two orders of magnitude, they cannot share a patch embedding~\cite{dosovitskiy2021vit}. The atmospheric stream uses a fixed patch size of 4, projecting each 20$\times$48 frame to $N = 5\times12 = 60$ tokens via a strided convolution. The ocean stream uses a dynamic patch embedding with patch size 64, keeping the token count tractable at 501$\times$1201 resolution and yielding $N_o = 8\times 19 = 152$ tokens per frame. The patch grid is computed at runtime from the input shape, so the same ocean encoder applies to differently sized domains without modification.

\subsection{Spatio-Temporal Stream Encoders}

Each stream factorizes attention into a spatial stage, mixing across the $N$ tokens per frame, and a temporal stage, mixing across frames per location~\cite{bertasius2021timesformer}. For the atmospheric stream,
\begin{equation}
    h_{a,\tau} = \mathrm{SpatialAttn}(X_{a,\tau}), \quad \tau = 1,\dots,28,
\end{equation}
\begin{equation}
    a = \mathrm{TemporalAttn}(h_{a,1},\dots,h_{a,28}),
\end{equation}
using 8 spatial layers then 4 temporal layers, with a learned frame embedding in place of a causal mask, since every frame in the window precedes the prediction time. The spatial summary $s$ is kept separate from the temporal summary $a$: $s$ answers where current forcing sits, $a$ how it evolved over the week, two physically distinct questions.

The ocean stream's spatial attention uses 2D Rotary Positional Embedding (RoPE)~\cite{su2024roformer,heo2024ropevit} rather than absolute position:
\begin{equation}
    h_{o,\tau} = \mathrm{SpatialAttn}_{\mathrm{RoPE}}(X_{o,\tau}), \quad \tau=1,\dots,60,
\end{equation}
\begin{equation}
    \tilde{o} = \mathrm{TemporalAttn}(h_{o,1},\dots,h_{o,60}),
\end{equation}
after which the ocean summary is pooled to match the atmospheric token count, $o = \mathrm{AdaptiveAvgPool}(\tilde{o}) \in \mathbb{R}^{N \times D}$, since fusion requires a common grid despite the ocean stream's finer resolution.

\begin{figure*}[htbp]
    \centering
    \includegraphics[width=.95\linewidth,trim=0 145 0 145,clip]{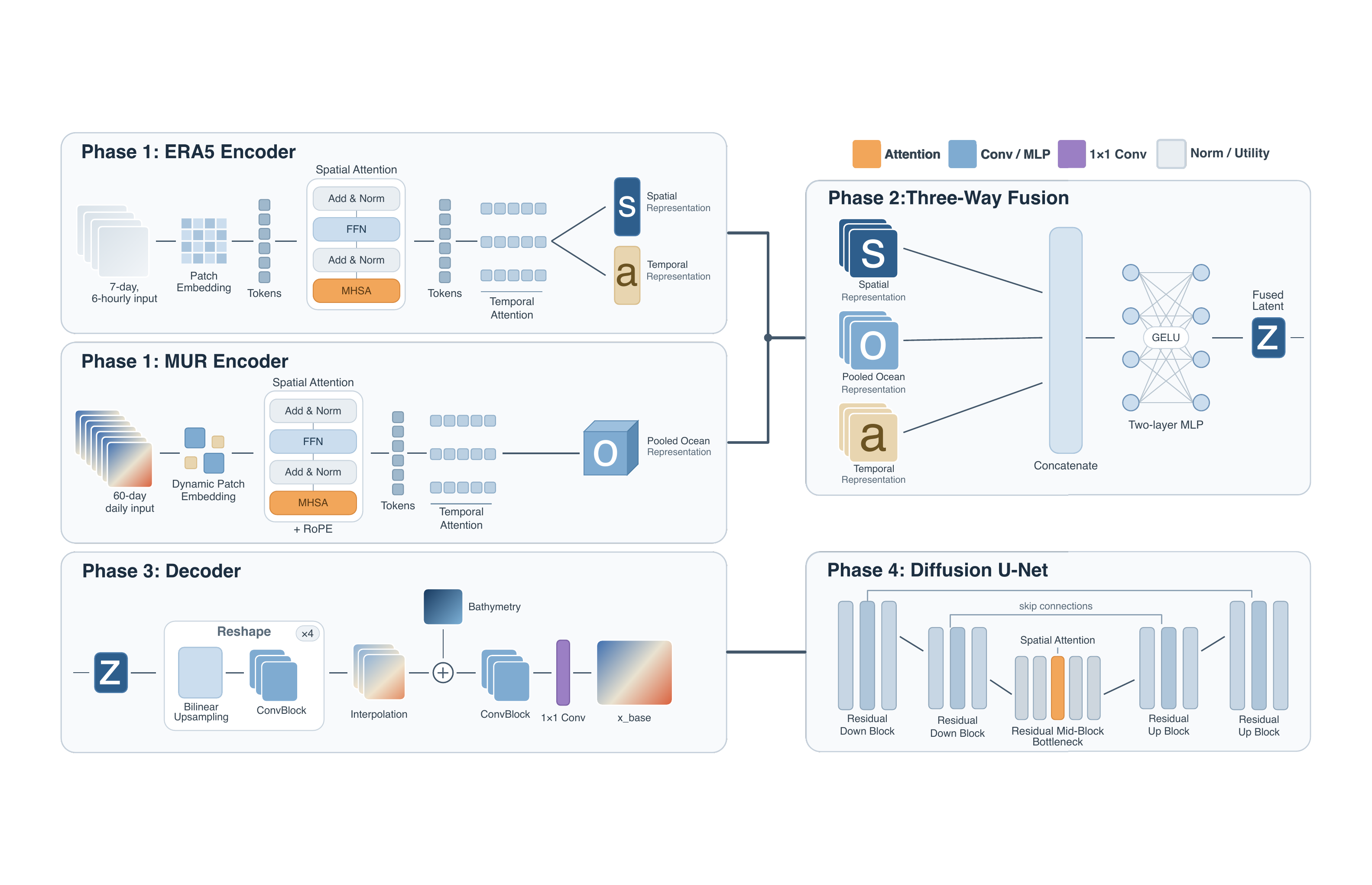}
    \caption{The \EddyFlow{} architecture. An atmospheric (ERA5) encoder processes a 7-day, 6-hourly forcing window and an ocean (MUR) encoder a 60-day daily SST history, each factorizing attention into spatial and temporal stages, with RoPE in the ocean stream. Spatial ($s$), temporal ($a$), and pooled ocean ($o$) summaries are fused by a two-layer MLP into a latent $z$, which the decoder upsamples under bathymetric guidance into the coarse tendency $x_{\mathrm{base}}$; a diffusion U-Net refines the fine-scale residual.}
    \label{fig:architecture}
\end{figure*}

\subsection{Three-Way Fusion}

The atmospheric spatial summary, temporal summary, and pooled ocean summary are concatenated and fused with a two-layer MLP rather than a single linear projection, since the mapping from three different summaries (instantaneous forcing location, forcing history, ocean memory) to a joint latent is unlikely to be linear:
\begin{equation}
    z = \mathrm{MLP}\big([\,s \,;\, a \,;\, o\,]\big) \in \mathbb{R}^{N \times D}.
\end{equation}

The concatenation keeps the spatial pattern $s$ distinct from the temporal trend $a$, since a forcing event's location and persistence differently constrain where and how much SST will change.

\subsection{Decoder and Bathymetric Guidance}

The latent $z$ is reshaped to $[D, n_H, n_W]$ and passed through a deterministic decoder producing a coarse baseline prediction of the SST tendency, $x_{\mathrm{base}} = R_\theta(z)$, via four bilinear upsampling stages. The output size is derived at runtime from the bathymetry tensor, so the same decoder targets different grid sizes without modification. Bathymetry is injected as a skip connection before the final convolution, $\mathrm{cat}[h, \mathrm{bathy}]$ with $h$ the penultimate feature map, so shelf breaks and banks shape the output where they are resolved, not only at the coarse encoder input where they modulate the forcing.

\subsection{Residual Formulation and Loss}
\label{sec:loss}

As formalized in Section~\ref{sec:problem}, \EddyFlow{} predicts the SST tendency $r_t = y_t - y_{t-1}$ rather than the absolute field and recovers $\hat{y}_t = y_{t-1} + \hat{r}_t$. The motivation is dynamic range. Absolute SST in the training domain spans tens of degrees Celsius, while the residual diffusion component models have a standard deviation nearly two orders of magnitude smaller. Learning the smaller signal concentrates capacity on the fine structure of daily change rather than the seasonal cycle, carried forward through $y_{t-1}$. The joint training objective, minimized in Phase~1 of the training schedule, is
\begin{equation}
    \mathcal{L} = \mathcal{L}_{\mathrm{rec}} + 10\,\mathcal{L}_{\mathrm{diff}} + 0.1\,\mathcal{L}_{\mathrm{spec}},
    \label{eq:total-loss}
\end{equation}
where $\mathcal{L}_{\mathrm{rec}}$ is the $L_1$ loss between the deterministic baseline $x_{\mathrm{base}}$ and the tendency $r_t$, $\mathcal{L}_{\mathrm{diff}}$ is the diffusion loss in Section~\ref{sec:edm}, and
\begin{equation}
    \mathcal{L}_{\mathrm{spec}} = \mathbb{E}\!\left[ w(k)\,\bigl|\,|\hat{Y}(k)| - |Y(k)|\,\bigr| \right], \quad w(k) = \sqrt{1 + 20k^2}.
    \label{eq:spec-loss}
\end{equation}

The weight $w(k)$ grows approximately linearly in wavenumber, penalizing missing high-wavenumber power more than low-wavenumber error, which counteracts the tendency of $\mathcal{L}_{\mathrm{rec}}$ to favor spatially smooth predictions. The weights $10$ and $0.1$ keep the initially small diffusion loss from being ignored and the large-for-textured-fields spectral loss from dominating reconstruction.

\subsection{Diffusion Refinement with DDIM}
\label{sec:edm}

The deterministic baseline $x_{\mathrm{base}}$, even with $\mathcal{L}_{\mathrm{spec}}$, still fails to represent mesoscale variance that is only partially determined by the coarse input, so a diffusion model refines the correction residual
\begin{equation}
    \delta_t = r_t - x_{\mathrm{base}},
\end{equation}
the part of the tendency the deterministic decoder fails to capture. We write $\delta$ for $\delta_t$ below, omitting the time index. We use a standard linear-schedule diffusion formulation with DDIM sampling, since the residual is modeled directly as a noise-prediction problem under a DDPM-style forward process.
The forward process uses a linear noise schedule $\beta_t \in [10^{-4}, 0.02]$ over $T_{\text{diff}}$ steps, with $\alpha_t = 1 - \beta_t$ and $\bar{\alpha}_t = \prod_{s \le t}\alpha_s$. For a clean residual $\delta$ and Gaussian noise $\epsilon \sim \mathcal{N}(0,I)$, the noised sample is
\begin{equation}
    r_t = \sqrt{\bar{\alpha}_t}\,\delta + \sqrt{1-\bar{\alpha}_t}\,\epsilon.
\end{equation}
The denoiser $\epsilon_\theta(r_t, x_{\text{base}}, z, \text{bathy}, t)$ is trained to recover the injected noise directly:
\begin{equation}
    \mathcal{L}_{\mathrm{diff}} = \mathbb{E}_{t,\epsilon}\left[\left\|\epsilon_\theta(r_t, x_{\mathrm{base}}, z, \mathrm{bathy}, t) - \epsilon\right\|^2\right].
\end{equation}
At inference, DDIM sampling proceeds deterministically with $\eta=0$ via the $x_0$-prediction form,
\begin{equation}
    \hat{\delta}_0 = \frac{r_t - \sqrt{1-\bar{\alpha}_t}\,\epsilon_\theta}{\sqrt{\bar{\alpha}_t}},
    \quad
    r_{t-1} = \sqrt{\bar{\alpha}_{t-1}}\,\hat{\delta}_0 + \sqrt{1-\bar{\alpha}_{t-1}}\,\epsilon_\theta,
\end{equation}
run over a strided subsequence of the training timesteps for fast sampling. As in the rest of the model, the full training objective combines baseline, diffusion, and spectral terms,
\begin{equation}
    \mathcal{L} = \mathcal{L}_{\mathrm{base}} + 10\,\mathcal{L}_{\mathrm{diff}} + 0.1\,\mathcal{L}_{\mathrm{spec}}.
\end{equation}

\subsection{RoPE and Domain Conditioning}
\label{sec:rope}

Absolute positional encoding binds a token's representation to its row and column in the training grid. Trained with absolute 2D sinusoidal position, the ocean stream learned associations between fixed grid coordinates and features, associations that become meaningless once the same coordinates index a different basin. We therefore replace absolute position with 2D RoPE in the ocean stream's spatial attention, which rotates query and key vectors by an angle determined by token position so that the attention score $q_i^\top k_j \;\propto\; \cos\big(\theta(\Delta u_{ij},\,\Delta v_{ij})\big)$ depends only on the \emph{relative} row--column displacement $(\Delta u_{ij},\,\Delta v_{ij})$ between tokens $i$ and $j$, never their absolute position. An eddy of a given radius therefore produces the same attention pattern wherever, and in whichever domain, it appears, a property absolute encoding cannot provide.

In parallel, the diffusion head is conditioned on the per-domain mean and standard deviation of the residual field. At training time, these statistics default to $(0,1)$; at zero-shot inference, the true domain statistics are substituted, letting the head rescale its output to an unseen basin without gradient-based adaptation. Together they target the two most basin-specific quantities, spatial relational structure and the field's statistical scale, leaving the atmospheric stream's absolute positions as the component not yet transfer-invariant.

\section{Experiments and Analysis}
\label{sec:experiments}

\subsection{Experimental Setup}

Models are implemented in PyTorch~\cite{paszke2019pytorch} and trained on Digital Research Alliance of Canada infrastructure. Each run uses 4 CPU cores and one 3g.40~GB Multi-Instance GPU (MIG) slice of an NVIDIA H100; one MIG slice per run, without distributed data parallelism. This provides enough memory for all components while keeping runs reproducible and easy to schedule on a shared cluster.

\subsubsection{Dataset and Preprocessing}

ERA5 is the fifth-generation European Centre for Medium-Range Weather Forecasts (ECMWF)\footnote{\url{https://www.ecmwf.int}} atmospheric reanalysis, combining short-range forecasts with observations through data assimilation into a physically consistent, globally complete hourly atmospheric state~\cite{era5}. We use ERA5 at 6-hourly resolution (00, 06, 12, 18 UTC), cropped to 47$^\circ$N--52$^\circ$N, 68$^\circ$W--56$^\circ$W, and regridded to a coarse 20$\times$48 grid at 0.25$^\circ$ spacing.

MUR (Multi-scale Ultra-high Resolution)\footnote{\url{https://podaac.jpl.nasa.gov/dataset/MUR-JPL-L4-GLOB-v4.1}} SST is a satellite-derived Level-4 product that fuses infrared and microwave observations with in-situ measurements into a gap-filled daily SST analysis at approximately 1~km resolution~\cite{mur}. We use MUR at its native 0.01$^\circ$ spacing over the same crop, giving a 501$\times$1201 fine grid. The training target is the SST tendency $r_t = y_t - y_{t-1}$ of Section~\ref{sec:problem}, computed from the MUR field; the absolute field is retained only to reconstruct $\hat{y}_t$ at evaluation. The training domain is the GSL; the zero-shot and few-shot evaluation domains are BOF and GOM. Data are split temporally, where training uses 2013--2020 (8 years), validation uses 2021, and test uses 2022--2023 (732 valid samples). Normalization statistics are computed over each stream's training-period data.

ERA5's SST channel is not an independently analyzed ocean product; it is a lower boundary condition prescribed from an external, coarser SST/sea-ice product used only to force the atmospheric model rather than to resolve ocean structure~\cite{era5sstboundary}. ERA5 SST is thus unsuited to fine-scale reconstruction and, in our crop, shows substantially weaker gradients than the MUR analysis.

\subsubsection{Evaluation Metrics}

We report the three metrics of Sections~\ref{sec:ss} and~\ref{sec:psdr}, instantiated as follows. First, RMSE is pixel-and-day-pooled over the full test period, restricted to valid ocean pixels only:
\begin{equation}
    \mathrm{RMSE} = \sqrt{\frac{\sum (\hat{y} - y)^2 \cdot m}{\sum m}},
\end{equation}
where $m$ is the valid mask, excluding pixels with sea-ice fraction above 0.15 or non-finite SST values. RMSE is computed in normalized units and converted to Celsius for reporting, except on the GOM, where figures and tables state normalized units and $^\circ$C values follow by multiplying by $\sigma = 6.3153$. Second, we report the skill score of Eq.~\eqref{eq:skill}, where the persistence reference is ERA5's coarse SST channel, denormalized and bilinearly upsampled to the fine grid; this is the sole persistence denominator used throughout the paper. Third, we report the PSD ratio of Section~\ref{sec:psdr}, the mean azimuthally averaged 2D PSD of prediction over target in the 5--50~km band, plus the logarithmic bias $\log_{10}(\mathrm{PSDR})$. Correlation is not a primary aggregate metric. All evaluations provide the model with ERA5 inputs up to time $t$ and MUR history up to time $t-1$.

\subsubsection{Models and Training Protocol}

\begin{table*}[t]
    \centering
    \caption{The six-stage architectural ladder used for all experiments. Stage~1, joint spatio-temporal attention with a residual head, serves as the strong downscaling baseline; each later stage modifies one part of the design (rightmost column), so metric changes between adjacent stages localize, though do not perfectly isolate, each component's contribution.}
    \label{tab:stage_ladder}
    \footnotesize
    \setlength{\tabcolsep}{6pt}
    \renewcommand{\arraystretch}{1.2}
    \begin{tabular}{@{}p{2.0cm}p{7.0cm}p{3.6cm}p{4.0cm}@{}}
        \toprule
        \textbf{Stage} & \textbf{Encoder / Representation}                   & \textbf{Diffusion Head} & \textbf{Key Addition}        \\
        \midrule
        Stage 1           & Joint spatio-temporal self-attn., block-causal mask & Baseline residual       & --- (ladder baseline)        \\
        Stage 2           & Alternating spatial/temporal blocks, single stream  & Baseline residual       & Factorized attention         \\
        Stage 3           & Dual-stream, cross-attn. bridges                    & Baseline residual       & Learned stream fusion        \\
        Stage 4 (EddyFlow)           & Separate atm./ocean streams, 2D RoPE (ocean)        & DDIM~\cite{song2021ddim} & RoPE + DDIM refinement       \\
        Stage 5           & Same as Stage 4 (shared encoder)                    & EDM~\cite{karras2022} + domain norm.      & Domain norm. for zero-shot   \\
        Stage 6           & Stage 4 encoder, ocean pathway via 8$\times$ VAE    & EDM (latent)            & Latent compression for scale \\
        \bottomrule
    \end{tabular}
\end{table*}

Table~\ref{tab:stage_ladder} defines the six-stage ladder used throughout the evaluation. Stage~1, a single joint spatio-temporal attention model with a residual head, serves as the strong downscaling baseline; each subsequent stage builds into \EddyFlow{}.

Training proceeds in three phases; Phase~1 jointly trains the encoder, decoder, and diffusion module for 100 epochs on the GSL under the objective of Eq.~\eqref{eq:total-loss}; Phase~2 fine-tunes the diffusion module (encoder and decoder frozen) for 40 epochs on the BOF and the GOM; and Phase~3 performs few-shot transfer for 20 epochs at 1-, 7-, and 30-day target windows. All zero-shot results use the Phase~1 checkpoint; Phases~2 and~3 constitute the few-shot experiments. All phases use AdamW~\cite{loshchilov2019adamw} ($\beta_1,\beta_2=0.9,0.95$, weight decay $10^{-4}$), with learning rates $3\times10^{-4}$ and $10^{-5}$ for joint training and fine-tuning under cosine annealing to $0.01\times$, bfloat16 mixed precision, gradient clipping at 1.0 post-accumulation, no early stopping, and non-finite-loss batches skipped rather than aborting.

\subsubsection{Baselines}

We report two baselines. The first is ERA5 coarse persistence, constructed as in the skill-score definition above; it serves as the skill-score denominator and represents no learned downscaling model beyond the reanalysis SST channel itself. The second is MUR-yesterday, which is reported only as an oracle upper-bound reference and is not used to compute the main skill score, since the MUR history is an input to the model (Section~\ref{sec:ss}). We include no separate classical interpolation or super-resolution baseline; bilinear up-sampling appears only inside the ERA5 persistence construction, not as an independent competitor.

\subsection{GSL Results}

Table~\ref{tab:gsl_summary} collects in-domain GSL performance across all six stages for RMSE, skill, and PSD ratio. Across all three metrics, performance is largely set by the early stages and, except for Stage~5's latent-domain variant, changes marginally thereafter, confirming that the ladder's later stages are not evaluated on further in-domain gains; the transfer results below should be read against this baseline.

\begin{table}[h]
    \centering
    \caption{In-domain GSL RMSE ($^\circ$C), persistence-relative skill, and 5--50~km PSD ratio for all six stages, together with the zero-shot PSD ratio on the transfer basins (BOF, GOM). PSD values closer to 1 indicate closer agreement with target spectral power. In-domain performance changes little across stages, so the ladder's differences emerge under transfer.}
    \label{tab:gsl_summary}
    \small
    \setlength{\tabcolsep}{6pt}
    \resizebox{\columnwidth}{!}{%
        \begin{tabular}{lccccc}
            \toprule
                    & \multicolumn{3}{c}{\textbf{Source domain (GSL)}} & \multicolumn{2}{c}{\textbf{Transfer PSD}}                            \\
            \cmidrule(lr){2-4} \cmidrule(lr){5-6}
            Model   & RMSE ($^{\circ}$C)                               & Skill                                     & PSD    & BOF    & GOM    \\
            \midrule
            Stage 1 & 0.5381                                           & 0.7340                                    & 1.0155 & 1.0258 & 0.9831 \\
            Stage 2 & 0.5494                                           & 0.7284                                    & 0.9897 & 1.0043 & 1.0092 \\
            Stage 3 & 0.5440                                           & 0.7311                                    & 0.9883 & 1.0022 & 0.9847 \\
            Stage 4 & 0.5522                                           & 0.7270                                    & 1.0019 & 0.9891 & 0.9738 \\
            Stage 5 & 0.6767                                           & 0.6637                                    & 1.0097 & 0.9996 & 1.0068 \\
            Stage 6 & 0.5751                                           & 0.7157                                    & 1.0000 & 1.0200 & 0.9603 \\
            \bottomrule
        \end{tabular}%
    }
\end{table}

\subsection{RMSE Results}

\begin{figure}[htbp]
    \centering
    \includegraphics[width=.95\linewidth]{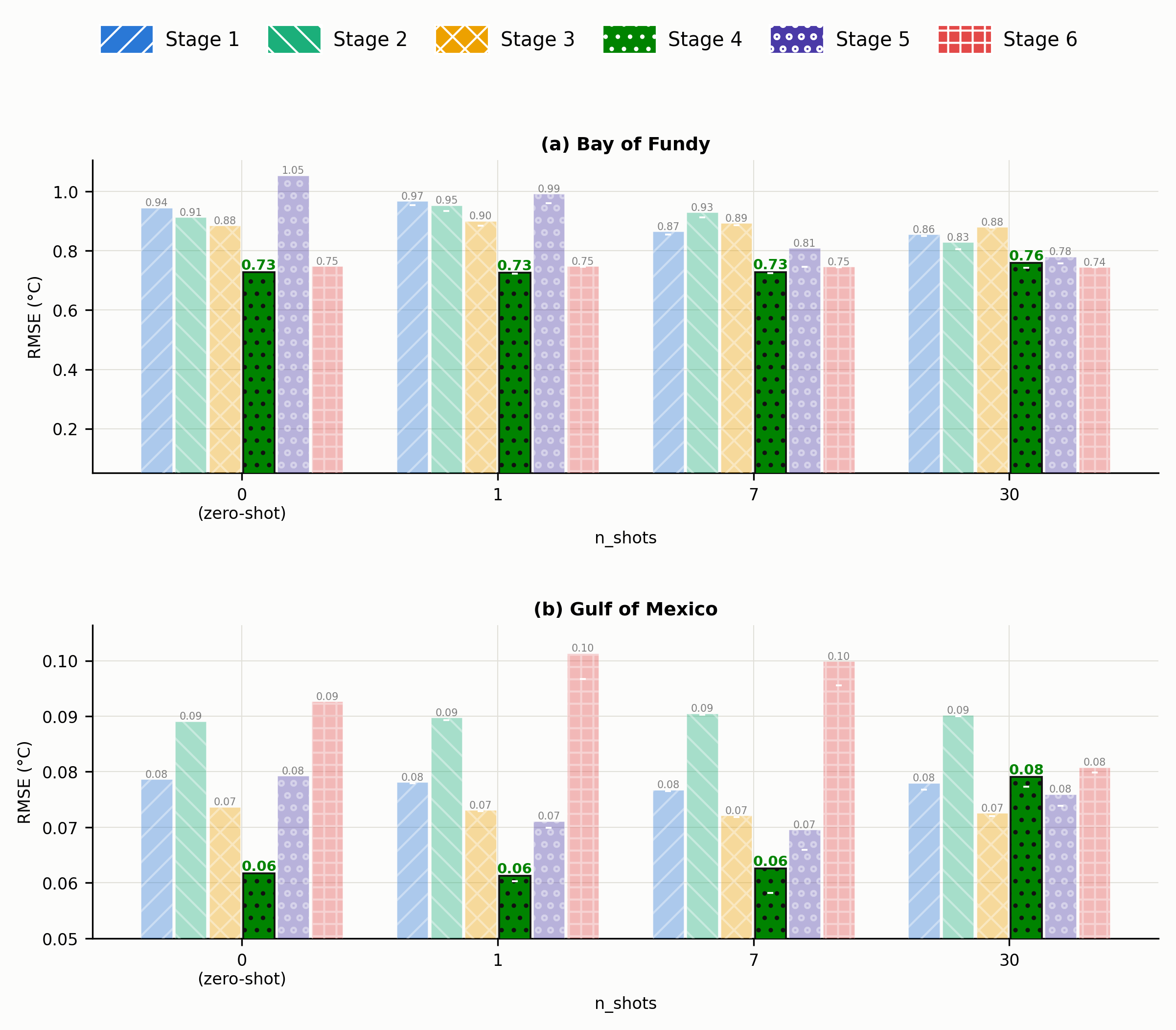}
    \caption{RMSE across the six-stage ladder on the (a) BOF and (b) GOM, versus the few-shot budget $n$, the number of target-domain days used for adaptation; lower is better.}
    \label{fig:rmse_results}
\end{figure}

Figure~\ref{fig:rmse_results} reports zero-shot RMSE across the six-stage ladder on GSL, BOF, and GOM. Source-domain error saturates almost immediately as, past the Stage~1$\to$2 transition, GSL RMSE stays within a narrow band, with Stage~5 as the sole outlier, so little of the ladder shows up in-domain. Transfer tells a different story as Stage~4 cuts BOF RMSE by roughly 22\% and GOM RMSE by roughly 21\% relative to the Stage~1 baseline, the largest transfer-domain gap in the ladder and the lowest zero-shot error on either transfer domain. With no GSL counterpart, the gap reflects better generalization from the representation, not better fitting of the training distribution.

Stages~5 and~6, despite more expressive generative refinement, do not improve on Stage~4's zero-shot RMSE; added diffusion-head flexibility does not translate into transfer accuracy. Few-shot adaptation adds little further evidence either way as Stage~4 moves by under 1\% at 1 day and is measurably worse by 30 days on both domains as the diffusion starts over-fitting, so the zero-shot checkpoint already sits near the transfer ceiling before any target supervision.

\subsection{Skill Score Results}

\begin{figure}[b]
    \centering
    \includegraphics[width=.95\linewidth]{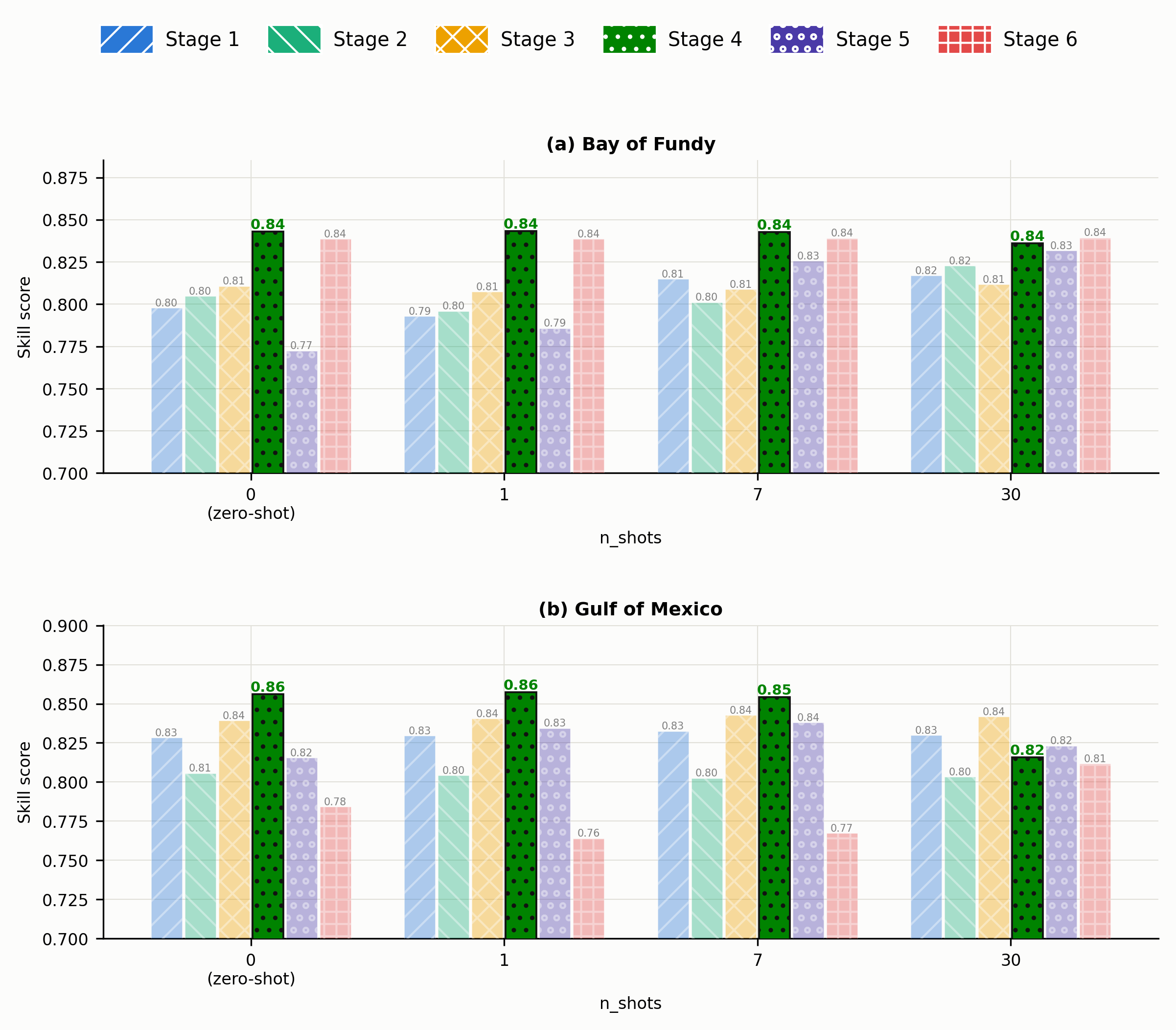}
    \caption{Skill relative to ERA5 persistence across the ladder stages on (a) BOF and (b) GOM versus the few-shot budget; higher is better; zero corresponds to carrying the reference forward, and positive values indicate improvement over it.}
    \label{fig:skill_results}
\end{figure}

Skill relative to ERA5 persistence (Figure~\ref{fig:skill_results}) reveals a pattern RMSE alone does not capture. Compared with Stage~1, Stage~2 leaves source-domain skill essentially unchanged, yet transfer skill fails to improve, dipping on GOM while staying flat on BOF, indicating that optimizing further for the GSL training distribution does not necessarily improve generalization. Stage~4 breaks this trend decisively as source-domain skill remains nearly unchanged, while transfer skill increases on both BOF and GOM. This matches the RMSE transition, now as an improvement over ERA5 persistence rather than absolute error.

Stage~5 reduces transfer skill relative to Stage~4, and Stage~6 remains competitive without consistently surpassing it; despite substantially different diffusion formulations, neither improves on Stage~4's transfer skill. This suggests that the primary source of the performance gain lies in the learned encoder representation rather than in the downstream diffusion refinement. Few-shot experiments reinforce this on both domains as the ranking barely changes after fine-tuning, so limited supervision mainly confirms the zero-shot representation rather than producing a better one.

\subsection{PSD Ratio Results}

PSD ratio in the 5--50~km band (Table~\ref{tab:gsl_summary}) answers a different question, asking not how large the error is but whether fine-scale structure survives. The ladder settles even earlier than in RMSE or skill. Stage~1 is the only model with a visible departure from spectral parity on GSL, and every later stage stays close to 1.0 across all three domains. That rules out the simplest alternative explanation as Stage~4 is not buying lower RMSE by smoothing away the variance PSD measures.

What makes this metric distinctive is how little it moves across Stages~4--6 even though their diffusion heads differ substantially and their RMSE and skill numbers disagree. A shared encoder producing near-identical spectral fidelity under three refinement strategies is the clearest evidence that preserved mesoscale content is a property of the representation, not the generative head. Few-shot fine-tuning moves this metric least as PSD stays near parity at every shot level, so adaptation recalibrates without altering retained spatial content.

\begin{figure*}[t]
    \centering
    \includegraphics[width=.87\linewidth,trim={0 0 8 0},clip]{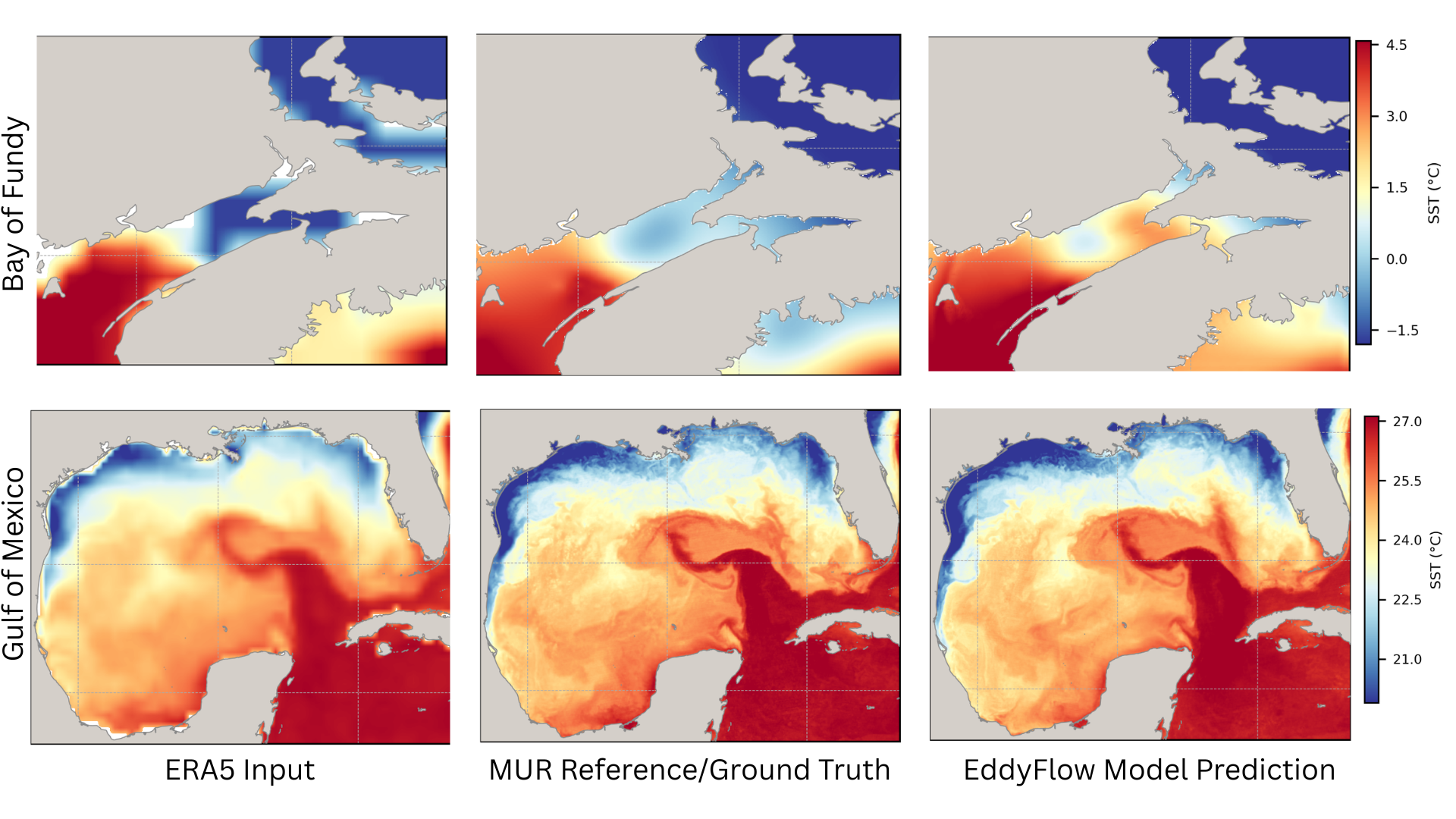}
    \caption{%
      Zero-shot downscaling under geographic domain shift for two basins held out of training. Each row shows a single date in an unseen basin: the Bay of Fundy (top) and the Gulf of Mexico (bottom), chosen to span contrasting thermal regimes. From left to right, the panels present the coarse input, the MUR ground-truth analysis, and the Stage~4 zero-shot prediction. The prediction is produced with no supervision from the target basin, showing that the learned representation reconstructs mesoscale sea-surface-temperature structure across basins rather than memorizing the training domain.
    }
    \label{fig:model_results}
\end{figure*}

Figure~\ref{fig:model_results} illustrates the qualitative counterpart of this transition. Together, the three metrics localize the gains in the encoder as Stage~4's dual-stream, RoPE-based representation enables zero-shot transfer across accuracy, skill, and spectral fidelity; Stages~5 and~6 mainly modulate adaptation and stability; and few-shot supervision leaves the ranking intact. Detailed ablations appear in the appendix.

\section{Conclusion}
\label{sec:conclusion}

This paper introduced \EddyFlow{}, a physics-informed framework for kilometer-scale SST downscaling. It encodes atmospheric forcing and ocean memory in separate streams at their native timescales, applies relative (RoPE) spatial encoding in the ocean pathway, injects bathymetry where it is resolved, and refines a deterministic tendency with a diffusion head under a wavenumber-weighted spectral loss, with performance judged jointly by RMSE, persistence-relative skill, and the PSD ratio. Trained on the GSL, it transfers zero-shot to the BOF and the GOM, cutting RMSE by up to 21\% over a strong baseline, reaching up to 85.6\% skill over ERA5 persistence, and holding the 5--50~km PSD ratio near 1.00. These gains localize in the dual-stream encoder, where RoPE-based representation sets transfer performance, while richer diffusion heads alter adaptation and stability without improving zero-shot accuracy, so transfer-aligned inductive bias matters more than added generative complexity.

Three limitations qualify these results. (i) the atmospheric stream still relies on absolute positions, so one pathway remains bound to training-domain geometry and plausibly contributes to the residual skill degradation observed under transfer; (ii) evaluation covers a single source basin and two transfer basins in the western North Atlantic, leaving eastern-boundary upwelling systems, tropical basins, and marginal ice zones untested, with ice-affected pixels masked rather than modeled; (iii) because the model predicts a one-day tendency from observed high-resolution history, multi-day horizons would require auto-regressive roll-out whose stability remains untested, and few-shot adaptation overfits by the 30-day window, eroding rather than extending the zero-shot advantage.

The work following these results should continue by extending relative positions to the atmospheric stream, training across multiple basins or through meta-learning to test whether generalization compounds, stabilizing auto-regressive roll-out for multi-day forecasting, and calibrating probabilistic downscaling from the stochastic head under spread--skill evaluation. Beyond methodology, coupling outputs to front-aware fisheries management and coastal hazard forecasting would test whether preserved mesoscale structure yields decision-relevant value, closing the loop between representation quality and the real-world use of the proposal.

\vspace{.3cm}\noindent
{\LARGE \textit{\textbf{Acknowledgments}}}
\vspace{.1cm}\\
This work was supported by the Natural Sciences and Engineering Research Council of Canada --- NSERC, the Faculty of Computer Science of Dalhousie University, and the Conselho Nacional de Desenvolvimento Cient\'{i}fico e Tecnol\'{o}gico --- CNPq, Brasil.

\vspace{.3cm}\noindent
{\LARGE \textit{\textbf{Data Availability}}}
\vspace{.1cm}\\
The ERA5 atmospheric reanalysis is publicly available from the Copernicus Climate Data Store, and the Multiscale Ultrahigh Resolution (MUR) sea-surface temperature analysis from the NASA JPL Physical Oceanography Distributed Active Archive Center. The code that reproduces pre-processing, training, evaluation, and figure generation is available at an anonymized repository for peer review (\url{https://anonymous.4open.science/r/EddyFlow}). Because every input product is publicly available, all reported forecasting, skill-score, and ablation results can be reproduced from open data.

\vspace{.3cm}\noindent
{\LARGE \textit{\textbf{Responsible AI Use}}}
\vspace{.1cm}\\
Generative AI technologies supported initial manuscript drafting and the creation of preliminary code scaffolding. The authors independently reviewed, verified, and revised all AI-assisted material and retain full responsibility for the study's methods, claims, analyses, interpretations, figures, and reported results.

\clearpage
\bibliographystyle{ACM-Reference-Format}
\bibliography{references}

\appendix
\vspace{.35cm}\noindent
\textit{\textbf{\LARGE Appendix}}

\section{List of Acronyms}
\begin{table}[h]
    \centering
    \caption{Acronyms used throughout the paper.}
    \label{tab:acronyms}
    \small
    \setlength{\tabcolsep}{6pt}
    \resizebox{\columnwidth}{!}{%
        \begin{tabular}{ll}
            \toprule
            Acronym        & Expansion                                          \\
            \midrule
            \textbf{SST}   & Sea Surface Temperature                            \\
            \textbf{RMSE}  & Root-Mean-Square Error                             \\
            \textbf{PSD}   & Power Spectral Density                             \\
            \textbf{OOD}   & Out-Of-Distribution                                \\
            \textbf{GSL}   & Gulf of St.~Lawrence (training domain)             \\
            \textbf{BOF}   & Bay of Fundy (transfer domain)                     \\
            \textbf{GOM}   & Gulf of Mexico (transfer domain)                   \\
            \textbf{ERA5}  & ECMWF Reanalysis v5                                \\
            \textbf{ECMWF} & European Centre for Medium-Range Weather Forecasts \\
            \textbf{MUR}   & Multi-scale Ultra-high Resolution (SST product)    \\
            \textbf{RoPE}  & Rotary Positional Embedding                        \\
            \textbf{EDM}   & Elucidated Diffusion Model                         \\
            \textbf{DDPM}  & Denoising Diffusion Probabilistic Model            \\
            \textbf{DDIM}  & Denoising Diffusion Implicit Model                 \\
            \textbf{VAE}   & Variational Autoencoder                            \\
            \textbf{MLP}   & Multi-Layer Perceptron                             \\
            \textbf{CNN}   & Convolutional Neural Network                       \\
            \textbf{MIG}   & Multi-Instance GPU                                 \\
            \bottomrule
        \end{tabular}%
    }
\end{table}

\section{List of Notations}
\begin{table}[h]
    \centering
    \caption{Mathematical notation used throughout the paper.}
    \label{tab:notation}
    \small
    \setlength{\tabcolsep}{6pt}
    \resizebox{\columnwidth}{!}{%
        \begin{tabular}{ll}
            \toprule
            \textbf{Symbol} & \textbf{Description} \\
            \midrule
            $t$ & Daily time index \\
            $x_t$ & Coarse-resolution atmospheric input at time $t$ \\
            $o_t$ & Ocean-state history input (Section~\ref{sec:problem}) \\
            $y_t,\ \hat{y}_t$ & Ground-truth and predicted high-resolution SST field \\
            $r_t,\ \hat{r}_t$ & SST tendency $y_t - y_{t-1}$ and its prediction \\
            $\hat{y}^{\mathrm{pers}}_t$ & Persistence predictor $y_{t-1}$ \\
            $x_{\mathrm{base}}$ & Deterministic decoder's tendency prediction \\
            $\delta_t$ & Correction residual $r_t - x_{\mathrm{base}}$ refined by diffusion (written $\delta$) \\
            $f_\theta,\ R_\theta,\ F_\theta$ & Full mapping, decoder, and diffusion network \\
            $\mathbf{u}=(u_1,u_2)$ & Spatial position \\
            $y'(\mathbf{u})$ & Mean-removed SST anomaly field \\
            $\mu_y$ & Spatial mean of the SST field \\
            $\mathbf{k}=(k_x,k_y)$ & Spatial wavenumber (Fourier frequency) \\
            $k$ & Radial wavenumber magnitude \\
            $Y(\mathbf{k}),\ \hat{Y}(\mathbf{k})$ & 2-D Fourier transforms of target and prediction \\
            $P_y(\mathbf{k})$ & 2-D power spectrum of $y$ \\
            $\bar{P}_y(k)$ & Radially-averaged (isotropic) power spectrum \\
            $\Omega_k$ & Set of wavenumbers in radial bin $k$ \\
            $\mathrm{PSDR}(k)$ & Power spectral density ratio (spectral fidelity) \\
            $\varepsilon$ & Small stabilizing constant in the PSDR (distinct from the noise sample $\epsilon$) \\
            $X_a,\ X_o$ & Atmospheric (ERA5) and oceanic input tensors \\
            $\tau$ & Temporal frame index \\
            $h_{a,\tau},\ h_{o,\tau}$ & Per-frame spatial embeddings \\
            $\tilde{o}$ & Ocean temporal summary before pooling \\
            $s,\ a,\ o$ & Atmospheric spatial, atmospheric temporal, and pooled ocean summaries \\
            $q_i,\ k_j$ & Query and key vectors of tokens $i$ and $j$ \\
            $(\Delta u_{ij},\,\Delta v_{ij})$ & Relative row--column displacement between tokens $i$ and $j$ \\
            $\theta(\cdot)$ & RoPE rotation angle as a function of the displacement \\
            $z$ & Fused conditioning representation ($\mathbb{R}^{N\times D}$) \\
            $N,\ D$ & Fused token count and embedding dimension \\
            $N_o$ & Ocean-stream token count per frame \\
            $n_H,\ n_W$ & Latent grid height and width at the decoder input \\
            $h$ & Decoder's penultimate feature map \\
            $\mathrm{bathy}$ & Static bathymetry tensor injected at the decoder skip \\
            $\mathcal{L}$ & Total training loss \\
            $\mathcal{L}_{\mathrm{rec}},\ \mathcal{L}_{\mathrm{diff}},\ \mathcal{L}_{\mathrm{spec}}$ & Reconstruction, diffusion, and spectral loss terms \\
            $w(k)$ & Spectral loss frequency weighting $\sqrt{1 + 20k^2}$ \\
            $D_\theta,\ v$ & Diffusion denoiser and its noisy input \\
            $\sigma,\ \sigma_{\mathrm{data}}$ & Noise level and data standard deviation \\
            $\gamma$ & EDM preconditioning normalizer $\sigma^2+\sigma_{\mathrm{data}}^2$ \\
            $c_{\mathrm{skip}},c_{\mathrm{out}},c_{\mathrm{in}}$ & EDM preconditioning coefficients \\
            $\lambda(\sigma)$ & Diffusion loss weighting \\
            $\epsilon$ & Standard Gaussian noise sample (distinct from the constant $\varepsilon$) \\
            $m$ & Valid-pixel (ocean) mask \\
            $n$ & Few-shot adaptation budget (days of target-domain data) \\
            $\mathrm{SS}$ & Persistence-relative skill score \\
            $\mu,\ \sigma$ & GSL training SST normalization mean and scale (unrelated to the noise level $\sigma$) \\
            $\dsym\mathrm{RMSE},\ \dsym\mathrm{Skill},\ \dsym\mathrm{PSD}$ & No-MUR ablation percentage deltas \\
            \bottomrule
        \end{tabular}%
    }
\end{table}

\section{Domain Characterization and OOD Analysis}

This appendix characterizes the three ocean domains used in the \EddyFlow{} evaluation: GSL (training domain), BOF, and GOM. It summarizes grid specifications, target SST distributions, ERA5 atmospheric input statistics, bathymetric properties, the shared normalization pipeline, and the main out-of-distribution (OOD) differences between the source domain and the two transfer domains.

\subsection{Domain overview}

GSL covers 44.0--52.8$^\circ$N and 69.5--56.0$^\circ$W and is the only Phase~1 training domain. It is a sub-polar estuary with a large seasonal SST cycle, winter sea ice, strong freshwater influence, and mixed shelf/deep-channel bathymetry~\cite{gsl_biogeochem_2021}. BOF covers 44.0--47.0$^\circ$N and 67.0--63.0$^\circ$W and is a zero-shot and few-shot evaluation domain. It overlaps GSL in latitude and has a very similar SST range, but its tidal forcing is much stronger. GOM covers 18.0--31.0$^\circ$N and 98.0--80.0$^\circ$W and is likewise used for zero-shot and few-shot evaluation. It is the most different domain, with much warmer SST, no ice, and circulation dominated by the Loop Current and warm-core eddies.

\subsection{Grid specifications}
The MUR SST target is evaluated on a 0.01$^\circ$ grid, which corresponds to roughly 1 km resolution. The fine-grid sizes are 500$\times$1200 for GSL, 301$\times$401 for BOF, and 1301$\times$1801 for GOM. ERA5 is provided at 0.25$^\circ$ resolution and is used as the coarse atmospheric input. The resulting downscaling factor is approximately 25$\times$ in each spatial direction for GSL and GOM, and about 23--24$\times$ for BOF.

\subsection{Target SST statistics}

Table~\ref{tab:domain_summary} summarizes the MUR SST target distribution for the 2022--2023 test period. GSL and BOF have similar SST means and ranges, whereas GOM is shifted to much higher temperatures and lies mostly outside the GSL training SST distribution.
\begin{table}[t]
    \centering
    \caption{Geography, MUR SST target statistics over the 2022--2023 test period, bathymetry, and OOD indicators for the three ocean basins; BOF and GOM SST values are converted to $^\circ$C before statistics are computed. The indicators quantify the transfer-difficulty ordering used in the analysis; BOF is a mild OOD case (full SST overlap with the GSL training range, no shifted ERA5 channels), while GOM is a strong one (26\% overlap, 14 of 21 channels shifted beyond 0.5$\sigma$).}
    \label{tab:domain_summary}
    \small
    \setlength{\tabcolsep}{6pt}
    \resizebox{\columnwidth}{!}{%
        \begin{tabular}{lccc}
            \toprule
            Property                              & GSL                 & BOF                 & GOM                 \\
            \midrule
            \addlinespace[2pt]
            \multicolumn{4}{@{}l}{\textbf{\textit{Geography}}}                                                      \\
            \addlinespace[2pt]
            Latitude range                        & 44.0--52.8$^\circ$N & 44.0--47.0$^\circ$N & 18.0--31.0$^\circ$N \\
            Longitude range                       & 69.5--56.0$^\circ$W & 67.0--63.0$^\circ$W & 98.0--80.0$^\circ$W \\
            \midrule
            \addlinespace[2pt]
            \multicolumn{4}{@{}l}{\textbf{\textit{MUR SST target statistics ($^\circ$C)}}}                          \\
            \addlinespace[2pt]
            Mean                                  & 6.68                & 5.61                & 25.96               \\
            Std                                   & 6.69                & 5.62                & 3.06                \\
            P1                                    & -1.80               & -1.80               & 15.94               \\
            P5                                    & -1.52               & -1.79               & 20.44               \\
            P95                                   & 18.16               & 15.87               & 29.97               \\
            P99                                   & 19.78               & 17.97               & 30.78               \\
            Min                                   & -1.80               & -1.80               & 6.94                \\
            Max                                   & 22.82               & 20.07               & 32.76               \\
            \midrule
            \addlinespace[2pt]
            \multicolumn{4}{@{}l}{\textbf{\textit{Bathymetry}}}                                                     \\
            \addlinespace[2pt]
            Fine grid                             & 500$\times$1200     & 301$\times$401      & 1301$\times$1801    \\
            Ocean fraction (\%)                   & 45.1                & 73                  & 76                  \\
            Mean ocean depth (m)                  & 175                 & 25                  & 800                 \\
            Maximum depth (m)                     & 761                 & 235                 & 3750                \\
            10th percentile depth (m)             & 31                  & 2                   & 30                  \\
            90th percentile depth (m)             & 375                 & 65                  & 2100                \\
            \midrule
            \addlinespace[2pt]
            \multicolumn{4}{@{}l}{\textbf{\textit{OOD indicators}}}                                                 \\
            \addlinespace[2pt]
            SST shift from GSL (sig)              & 0.00                & -0.05               & +3.18               \\
            SST overlap with GSL range            & 100\%               & 100\%               & 26\%                \\
            ERA5 channels OOD $>0.5\sigma$        & 0/21                & 0/21                & 14/21               \\
            Sea ice present                       & Yes                 & Rarely              & Never               \\
            ERA5 persistence RMSE ($^\circ$C)     & 2.0229              & 4.6698              & 2.8924              \\
            MUR-yesterday oracle RMSE ($^\circ$C) & 0.5202              & 0.6818              & 0.3031              \\
            \bottomrule
        \end{tabular}%
    }
\end{table}
The shared SST normalization uses the GSL training mean and standard deviation, $\mu = 5.9023\,^\circ\mathrm{C}$ and $\sigma = 6.3153\,^\circ\mathrm{C}$, for all domains. Under this normalization, BOF lies almost entirely within the GSL training SST range, whereas GOM lies predominantly above it.

\subsection{ERA5 input statistics}
ERA5 atmospheric channels are normalized using per-channel mean and standard deviation computed on the GSL training period. The same statistics are applied to BOF and GOM without domain-specific rescaling. Table~\ref{tab:era5_stats} shows the GSL training statistics together with the corresponding test-period percentiles. The GSL training and test distributions are close, suggesting that degradation on BOF and GOM is not due to a shift within the training domain.
\begin{table}[t]
    \centering
    \caption{ERA5 per-channel normalization statistics computed on the GSL training period (2013--2020), applied unchanged to BOF and GOM, alongside GSL test-period 1st and 99th percentiles. The proximity of train and test distributions indicates that transfer degradation reflects cross-basin shift rather than temporal drift within the source domain.}
    \label{tab:era5_stats}
    \small
    \setlength{\tabcolsep}{4pt}
    \resizebox{\columnwidth}{!}{%
        \begin{tabular}{lrrrrr}
            \toprule
            Channel   & Units       & Train mean  & Train std & Test P1    & Test P99    \\
            \midrule
            u10       & m/s         & 1.4096      & 4.4805    & -10.1506   & 11.9952     \\
            v10       & m/s         & -0.1007     & 4.3195    & -10.9579   & 10.8044     \\
            msl       & Pa          & 101298.4766 & 1030.9015 & 98790.7500 & 103437.1250 \\
            sst\_era5 & K           & 279.0615    & 6.3021    & 271.4602   & 292.8271    \\
            t2m       & K           & 276.4195    & 10.4942   & 248.4009   & 295.6465    \\
            siconc    & 0--1        & 0.0578      & 0.1708    & 0.0000     & 0.6984      \\
            u850      & m/s         & 5.0713      & 8.5292    & -18.9237   & 21.7731     \\
            v850      & m/s         & -0.1896     & 8.6014    & -19.6215   & 24.0440     \\
            t850      & K           & 271.4302    & 10.1253   & 248.6821   & 289.5664    \\
            z850      & m$^2$/s$^2$ & 13834.3350  & 930.4116  & 11629.2227 & 15505.4688  \\
            q850      & kg/kg       & 0.0034      & 0.0027    & 0.0002     & 0.0110      \\
            u700      & m/s         & 9.0247      & 9.3201    & -15.6331   & 29.0035     \\
            v700      & m/s         & 0.3513      & 9.3140    & -19.4134   & 26.0006     \\
            t700      & K           & 265.3427    & 9.0362    & 243.8658   & 281.1182    \\
            z700      & m$^2$/s$^2$ & 28823.8926  & 1301.7219 & 26069.1836 & 31150.6797  \\
            q700      & kg/kg       & 0.0020      & 0.0018    & 0.0001     & 0.0071      \\
            u500      & m/s         & 14.9958     & 13.0255   & -17.9777   & 46.5566     \\
            v500      & m/s         & 1.5033      & 12.9020   & -27.6660   & 34.7296     \\
            t500      & K           & 251.3449    & 8.4190    & 231.6444   & 266.2297    \\
            z500      & m$^2$/s$^2$ & 53825.8789  & 2039.6470 & 49469.4062 & 57373.8398  \\
            q500      & kg/kg       & 0.0007      & 0.0010    & 0.0000     & 0.0033      \\
            \bottomrule
        \end{tabular}%
    }
\end{table}
\subsection{Bathymetry}

Bathymetry is a static auxiliary input coarsened to the ERA5 grid, transformed with $\log(1+x)$, and normalized with GSL training statistics before encoder input. The GSL bathymetry contains the Laurentian Channel and a mixed shelf/deep-channel structure; BOF is much shallower, and GOM contains a much deeper open-ocean basin. These differences matter because the model uses bathymetry as a spatial prior for where fine-scale SST structure appears.

\subsection{Normalization pipeline}

All three domains use the same normalization statistics from the GSL 2013--2020 training period. No domain-specific rescaling is applied in the encoder or baseline decoder. BOF and GOM are converted from Kelvin to degrees Celsius before the shared SST normalization is applied. Where domain conditioning is enabled (Section~\ref{sec:rope}; Stage~5 in Table~\ref{tab:stage_ladder}), the diffusion head additionally injects the target-domain SST mean and standard deviation.

The SST target is converted to a z-score using the GSL training mean and standard deviation. Physical RMSE is recovered from normalized RMSE by multiplying by $\sigma=6.3153\,{^\circ C}$. Because the models predict SST changes rather than absolute SST levels, the constant GOM offset does not directly inflate the error; the day-to-day target is centered near zero in all domains.

\subsection{OOD analysis}

The OOD gap is small for BOF and large for GOM. On SST, BOF sits inside the GSL training range, while GOM lies mostly above it. On ERA5 forcing, BOF stays close to the GSL training distribution, whereas GOM shifts across many channels, especially moisture, temperature, and wind structure. In bathymetry, BOF is shallow and coastal, whereas GOM features a deeper open-ocean regime.

Table~\ref{tab:domain_summary} summarizes the main domain-level differences. BOF is a mild OOD case, changing the local tidal and bathymetric context but not the broad SST scale. GOM is a strong OOD case, changing SST level, atmospheric forcing regime, and circulation structure.

\section{Ablations}

We report two ablations: a hyperparameter and seed sweep summarized through the few-shot transfer experiments, and an input ablation that removes the MUR history stream (Appendix~\ref{app:no_mur_ablation}). The few-shot sweep covers all six stages, both transfer domains (BOF and GOM), three target-data budgets (1-, 7-, and 30-day windows), and three random seeds (0, 42, 123), for 108 configurations; Tables~\ref{tab:fewshot_bof} and~\ref{tab:fewshot_gom} report the best and mean result per stage and budget.

\begin{table*}[t]
    \centering
    \caption{Few-shot transfer results on the BOF across 1-, 7-, and 30-day adaptation budgets, reporting RMSE ($^\circ$C), skill score, and PSD ratio. Best and Mean are computed over three random seeds (0, 42, 123): Best is the minimum for RMSE, the maximum for Skill, and the value closest to 1.0 for PSD. The ERA5 persistence baseline RMSE on BOF is $\approx$~4.67~$^\circ$C (Table~\ref{tab:domain_summary}). Stage~5 (EDM diffusion) shows high seed variance at the 7-day budget ($\sigma_\text{RMSE}=0.073$~$^\circ$C) due to bimodal convergence dynamics.}
    \label{tab:fewshot_bof}
    \small
    \setlength{\tabcolsep}{4pt}
    \begin{tabular}{lcccccccccccccccccc}
        \toprule
                & \multicolumn{6}{c}{\textbf{1-day}} & \multicolumn{6}{c}{\textbf{7-day}} & \multicolumn{6}{c}{\textbf{30-day}}                                                                                                                                                                                                                   \\
        \cmidrule(lr){2-7} \cmidrule(lr){8-13} \cmidrule(lr){14-19}
        Model
                & \multicolumn{2}{c}{\textit{RMSE}}
                & \multicolumn{2}{c}{\textit{Skill}}
                & \multicolumn{2}{c}{\textit{PSD}}
                & \multicolumn{2}{c}{\textit{RMSE}}
                & \multicolumn{2}{c}{\textit{Skill}}
                & \multicolumn{2}{c}{\textit{PSD}}
                & \multicolumn{2}{c}{\textit{RMSE}}
                & \multicolumn{2}{c}{\textit{Skill}}
                & \multicolumn{2}{c}{\textit{PSD}}                                                                                                                                                                                                                                                                                                \\
        \cmidrule(lr){2-3} \cmidrule(lr){4-5} \cmidrule(lr){6-7}
        \cmidrule(lr){8-9} \cmidrule(lr){10-11} \cmidrule(lr){12-13}
        \cmidrule(lr){14-15} \cmidrule(lr){16-17} \cmidrule(lr){18-19}
                & Best                               & Mean                               & Best                                & Mean           & Best  & Mean
                & Best                               & Mean                               & Best                                & Mean           & Best  & Mean
                & Best                               & Mean                               & Best                                & Mean           & Best  & Mean                                                                                                                                                                                   \\
        \midrule
        Stage 1 & 0.955                              & 0.967                              & 0.796                               & 0.793          & 1.018 & 1.024 & 0.855          & 0.865          & 0.817          & 0.815          & 1.016 & 1.020 & 0.850          & 0.855          & 0.818          & 0.817          & 1.013          & 1.014 \\
        Stage 2 & 0.935                              & 0.953                              & 0.800                               & 0.796          & 1.003 & 1.007 & 0.914          & 0.929          & 0.804          & 0.801          & 1.001 & 1.007 & 0.806          & 0.828          & 0.828          & 0.823          & 1.009          & 1.015 \\
        Stage 3 & 0.886                              & 0.900                              & 0.810                               & 0.807          & 1.003 & 1.009 & 0.886          & 0.893          & 0.810          & 0.809          & 1.007 & 1.010 & 0.878          & 0.880          & 0.812          & 0.812          & 1.004          & 1.005 \\
        Stage 4 & \textbf{0.723}                     & \textbf{0.726}                     & \textbf{0.844}                      & \textbf{0.843} & 0.990 & 0.990 & \textbf{0.725} & \textbf{0.728} & \textbf{0.843} & \textbf{0.843} & 0.989 & 0.986 & \textbf{0.743} & 0.759          & \textbf{0.840} & 0.836          & 0.980          & 0.979 \\
        Stage 5 & 0.960                              & 0.992                              & 0.793                               & 0.786          & 1.006 & 1.021 & 0.746          & 0.808          & 0.839          & 0.826          & 1.004 & 1.009 & 0.758          & \textbf{0.780} & 0.836          & 0.832          & \textbf{0.999} & 1.001 \\
        Stage 6 & 0.747                              & 0.747                              & 0.839                               & 0.839          & 1.024 & 1.025 & 0.744          & 0.745          & 0.839          & 0.839          & 1.019 & 1.023 & 0.744          & 0.744          & 0.839          & \textbf{0.839} & 1.019          & 1.021 \\
        \bottomrule
    \end{tabular}
\end{table*}

\begin{table*}[t]
    \centering
    \caption{Few-shot transfer results on the GOM under the same protocol and seed conventions as Table~\ref{tab:fewshot_bof}. RMSE is reported in normalized units. Stage~4 leads at the 1- and 7-day budgets, while by 30 days Stage~3 attains the best RMSE and skill and Stage~5 reaches spectral parity.}
    \label{tab:fewshot_gom}
    \small
    \setlength{\tabcolsep}{4pt}
    \begin{tabular}{lcccccccccccccccccc}
        \toprule
                & \multicolumn{6}{c}{\textbf{1-day}} & \multicolumn{6}{c}{\textbf{7-day}} & \multicolumn{6}{c}{\textbf{30-day}}                                                                                                                                                                      \\
        \cmidrule(lr){2-7} \cmidrule(lr){8-13} \cmidrule(lr){14-19}
        Model
                & \multicolumn{2}{c}{\textit{RMSE}}
                & \multicolumn{2}{c}{\textit{Skill}}
                & \multicolumn{2}{c}{\textit{PSD}}
                & \multicolumn{2}{c}{\textit{RMSE}}
                & \multicolumn{2}{c}{\textit{Skill}}
                & \multicolumn{2}{c}{\textit{PSD}}
                & \multicolumn{2}{c}{\textit{RMSE}}
                & \multicolumn{2}{c}{\textit{Skill}}
                & \multicolumn{2}{c}{\textit{PSD}}                                                                                                                                                                                                                                                   \\
        \cmidrule(lr){2-3} \cmidrule(lr){4-5} \cmidrule(lr){6-7}
        \cmidrule(lr){8-9} \cmidrule(lr){10-11} \cmidrule(lr){12-13}
        \cmidrule(lr){14-15} \cmidrule(lr){16-17} \cmidrule(lr){18-19}
                & Best                               & Mean                               & Best                                & Mean           & Best  & Mean
                & Best                               & Mean                               & Best                                & Mean           & Best  & Mean
                & Best                               & Mean                               & Best                                & Mean           & Best  & Mean                                                                                                                                      \\
        \midrule
        Stage 1 & 0.078                              & 0.078                              & 0.830                               & 0.829          & 0.984 & 0.984 & 0.077          & 0.077 & 0.833          & 0.833 & 0.988 & 0.988 & 0.077          & 0.078 & 0.832 & 0.830 & 0.988          & 0.985 \\
        Stage 2 & 0.089                              & 0.090                              & 0.805                               & 0.804          & 1.010 & 1.010 & 0.090          & 0.091 & 0.803          & 0.802 & 1.010 & 1.010 & 0.090          & 0.090 & 0.804 & 0.803 & 1.010          & 1.010 \\
        Stage 3 & 0.073                              & 0.073                              & 0.841                               & 0.841          & 0.987 & 0.987 & 0.072          & 0.072 & 0.843          & 0.843 & 0.990 & 0.989 & 0.072          & 0.073 & 0.843 & 0.842 & 0.989          & 0.988 \\
        Stage 4 & \textbf{0.060}                     & \textbf{0.061}                     & \textbf{0.859}                      & \textbf{0.857} & 0.976 & 0.975 & \textbf{0.058} & 0.063 & \textbf{0.864} & 0.854 & 0.980 & 0.977 & 0.077          & 0.079 & 0.820 & 0.816 & 0.976          & 0.971 \\
        Stage 5 & 0.070                              & 0.071                              & 0.837                               & 0.834          & 0.992 & 0.990 & 0.066          & 0.070 & 0.846          & 0.838 & 0.996 & 0.992 & \textbf{0.074} & 0.076 & 0.828 & 0.823 & \textbf{1.000} & 1.000 \\
        Stage 6 & 0.097                              & 0.101                              & 0.775                               & 0.764          & 0.958 & 0.956 & 0.096          & 0.100 & 0.777          & 0.767 & 0.969 & 0.966 & 0.080          & 0.081 & 0.814 & 0.812 & 0.979          & 0.979 \\
        \bottomrule
    \end{tabular}
\end{table*}

\begin{table*}[t]
    \centering
    \caption{Percentage deltas for the no-MUR ablation, in which the 60-day MUR history is replaced by the z-scored climatological average; positive \(\dsym\)RMSE and \(\dsym\)Skill indicate degradation when MUR is removed, and \(\dsym\)PSD reports its signed relative change (positive when removal raises it). Results average seeds $\{0,42,123\}$; zero-shot uses a single checkpoint. Deltas are small almost everywhere, isolating Stage~5 on GOM at zero shot (+34.2\% RMSE, bold) as the only configuration with MUR dependence.}
    \label{tab:no_mur_delta_summary}
    \footnotesize
    \setlength{\tabcolsep}{2.1pt}
    \begin{tabular}{lcccccccccccccccccccccccc}
        \toprule
                                           & \multicolumn{6}{c}{\textbf{Zero-shot}} &
        \multicolumn{6}{c}{\textbf{1-day}} &
        \multicolumn{6}{c}{\textbf{7-day}} &
        \multicolumn{6}{c}{\textbf{30-day}}                                                                                                        \\
        \cmidrule(lr){2-7} \cmidrule(lr){8-13} \cmidrule(lr){14-19} \cmidrule(lr){20-25}
        Stage \& Domain
                                           & \multicolumn{2}{c}{$\dsym$RMSE}        &
        \multicolumn{2}{c}{$\dsym$Skill}   &
        \multicolumn{2}{c}{$\dsym$PSD}
                                           & \multicolumn{2}{c}{$\dsym$RMSE}        &
        \multicolumn{2}{c}{$\dsym$Skill}   &
        \multicolumn{2}{c}{$\dsym$PSD}
                                           & \multicolumn{2}{c}{$\dsym$RMSE}        &
        \multicolumn{2}{c}{$\dsym$Skill}   &
        \multicolumn{2}{c}{$\dsym$PSD}
                                           & \multicolumn{2}{c}{$\dsym$RMSE}        &
        \multicolumn{2}{c}{$\dsym$Skill}   &
        \multicolumn{2}{c}{$\dsym$PSD}                                                                                                             \\
        \cmidrule(lr){2-3} \cmidrule(lr){4-5} \cmidrule(lr){6-7}
        \cmidrule(lr){8-9} \cmidrule(lr){10-11} \cmidrule(lr){12-13}
        \cmidrule(lr){14-15} \cmidrule(lr){16-17} \cmidrule(lr){18-19}
        \cmidrule(lr){20-21} \cmidrule(lr){22-23} \cmidrule(lr){24-25}
                                           & Best                                   & Mean           & Best          & Mean          & Best & Mean
                                           & Best                                   & Mean           & Best          & Mean          & Best & Mean
                                           & Best                                   & Mean           & Best          & Mean          & Best & Mean
                                           & Best                                   & Mean           & Best          & Mean          & Best & Mean \\
        \midrule
        Stage 4, BOF                       & +3.0                                   & +3.0           & +0.6          & +0.6          & -0.5 & -0.5
                                           & +11.4                                  & +11.4          & +2.1          & +2.1          & -1.2 & -1.2
                                           & +3.2                                   & +3.2           & +0.6          & +0.6          & -0.7 & -0.7
                                           & +5.3                                   & +5.3           & +1.1          & +1.1          & -0.6 & -0.6 \\
        Stage 4, GOM                       & -1.6                                   & -1.6           & -0.2          & -0.2          & -0.1 & -0.1
                                           & +0.0                                   & +0.0           & -0.2          & -0.2          & -0.1 & -0.1
                                           & +0.0                                   & +0.0           & +0.0          & +0.0          & -0.6 & -0.6
                                           & -15.2                                  & -15.2          & -3.4          & -3.4          & -1.0 & -1.0 \\
        \midrule
        Stage 5, BOF                       & -1.5                                   & -1.5           & -0.4          & -0.4          & -1.3 & -1.3
                                           & +3.8                                   & +3.8           & +1.0          & +1.0          & +0.8 & +0.8
                                           & +1.5                                   & +1.5           & +0.4          & +0.4          & +0.2 & +0.2
                                           & +1.7                                   & +1.7           & +0.4          & +0.4          & -0.2 & -0.2 \\
        Stage 5, GOM                       & \textbf{+34.2}                         & \textbf{+34.2} & \textbf{+7.6} & \textbf{+7.6} & -1.9 & -1.9
                                           & +0.0                                   & +0.0           & -0.1          & -0.1          & +0.7 & +0.7
                                           & +0.0                                   & +0.0           & +0.0          & +0.0          & +0.6 & +0.6
                                           & +2.6                                   & +2.6           & +0.5          & +0.5          & +0.9 & +0.9 \\
        \midrule
        Stage 6, BOF                       & -1.1                                   & -1.1           & -0.1          & -0.1          & +0.3 & +0.3
                                           & -0.9                                   & -0.9           & -0.2          & -0.2          & +0.3 & +0.3
                                           & -0.9                                   & -0.9           & -0.2          & -0.2          & +0.3 & +0.3
                                           & -0.9                                   & -0.9           & -0.2          & -0.2          & +0.3 & +0.3 \\
        Stage 6, GOM                       & -2.2                                   & -2.2           & -0.4          & -0.4          & +0.0 & +0.0
                                           & -1.0                                   & -1.0           & -0.4          & -0.4          & +0.0 & +0.0
                                           & -2.0                                   & -2.0           & -0.5          & -0.5          & +0.0 & +0.0
                                           & -1.2                                   & -1.2           & -0.4          & -0.4          & +0.0 & +0.0 \\
        \bottomrule
    \end{tabular}
\end{table*}

\subsection{Evaluations Without MUR History}
\label{app:no_mur_ablation}

We assess how much the ocean stream depends on real target-domain MUR SST history by comparing the standard setup (with MUR) to a no-MUR variant in which the 60-day MUR input is replaced by the z-scored climatological average. All other factors (checkpoints, evaluation dates, seeds) are held fixed. Table~\ref{tab:no_mur_delta_summary} reports percentage deltas, oriented so that positive $\dsym$RMSE and $\dsym$Skill always indicate degradation when MUR is removed:
\begin{gather*}
    \dsym \mathrm{RMSE}\ (\%) = 100 \times \frac{\mathrm{RMSE}_{\text{no}} - \mathrm{RMSE}_{\text{with}}}{\mathrm{RMSE}_{\text{with}}},\\
    \dsym \mathrm{Skill}\ (\%) = 100 \times \frac{\mathrm{Skill}_{\text{with}} - \mathrm{Skill}_{\text{no}}}{\mathrm{Skill}_{\text{with}}},
\end{gather*}
while $\dsym \mathrm{PSD}\ (\%) = 100 \times (\mathrm{PSDR}_{\text{no}} - \mathrm{PSDR}_{\text{with}})/\mathrm{PSDR}_{\text{with}}$ reports its signed relative change, positive when removal raises it.

Across most stages, domains, and shot counts, these percentage deltas are small. RMSE changes are typically under 3\%, and skill changes lie within $\pm 1$\% for the majority of configurations. PSD ratios remain close to their original values, with most $|\dsym \text{PSD}|$ below 2\%, indicating that spectral fidelity is largely preserved without real MUR history. Stage 6 is the most robust as RMSE and skill deltas are consistently near zero or slightly favorable (up to $\approx 1$\% improvement in skill), and PSD changes are negligible. Stage 4 shows mild sensitivity on BOF at $n=1$ (around 2--3\% skill drop) but is essentially neutral or slightly beneficial on GOM, especially at $n=30$, where skill improves by about 3\% without MUR.

The clearest case of genuine MUR dependence appears for \textbf{Stage 5 on GOM in the zero-shot setting}, where removing MUR increases RMSE by roughly 34\% and reduces skill by about 8\%. This is the largest degradation in the ablation and stands out against the otherwise mild pattern. In few-shot, Stage 5 on GOM recovers by $n=7$--$30$, RMSE and skill deltas shrink to within $\pm 3$\%, and PSD changes remain small. Few-shot adaptation thus mitigates most sensitivity as $n$ increases from 1 to 30, skill and RMSE deltas contract, and PSD ratios stay stable. Overall, real target-domain MUR history is not uniformly essential; the framework is robust to its removal, with Stage 5 on GOM (zero-shot) as the primary exception.

\balance
\end{document}